\documentclass[a4paper,conference]{IEEEtran}
\def \OURS {$\mathcal{RMS}$-Net}

\ifCLASSINFOpdf
\else
\fi

\usepackage{amssymb}
\usepackage{graphicx}
\usepackage{amsmath}
\usepackage{booktabs}
\usepackage{caption}
\usepackage{bbm}
\usepackage{bm}
\usepackage{dsfont}
\usepackage{balance}

\newcommand{\tit}[1]{\smallbreak\noindent\textbf{#1.}}

\def \ie {\emph{i.e.}}
\def \eg {\emph{e.g.}}
\def \etal {\emph{et al.}}

\begin{document}
%
\title{\OURS:\\Regression and Masking for Soccer Event Spotting}

\author{\IEEEauthorblockN{Matteo Tomei$^1$, Lorenzo Baraldi$^1$, Simone Calderara$^1$, Simone Bronzin$^2$, Rita Cucchiara$^1$}
\IEEEauthorblockA{$^1$University of Modena and Reggio Emilia, $^2$Metaliquid S.R.L.\\
Email: $^1$\{name.surname\}@unimore.it, $^2$\{name.surname\}@meta-liquid.com}
}


%


\maketitle

\begin{abstract}
The recently proposed action spotting task consists in finding the exact timestamp in which an event occurs. This task fits particularly well for soccer videos, where events correspond to salient actions strictly defined by soccer rules (a \textit{goal} occurs when the ball crosses the goal line). In this paper, we devise a lightweight and modular network for action spotting, which can simultaneously predict the event label and its temporal offset using the same underlying features. We enrich our model with two training strategies: the first one for data balancing and uniform sampling, the second for masking ambiguous frames and keeping the most discriminative visual cues. When tested on the SoccerNet dataset and using standard features, our full proposal exceeds the current state of the art by 3 Average-mAP points. Additionally, it reaches a gain of more than 10 Average-mAP points on the test set when fine-tuned in combination with a strong 2D backbone.
\end{abstract}


%
\IEEEpeerreviewmaketitle

\section{Introduction}

Understanding videos has been one of the most attractive and challenging areas of Computer Vision of the last few years. While a significant research effort has been carried out to design novel architectures and training approaches for enhancing the effectiveness of spatio-temporal features extraction~\cite{feichtenhofer2019slowfast, Wu_2020_CVPR}, there is also a growing interest in bringing these techniques to more application-oriented domains~\cite{giancola2018soccernet, bertini2004semantic}. Among these, the soccer industry could greatly benefit from the automatic understanding of soccer matches. Soccer videos are indeed used by professionals for statistics generation, for analyzing and developing strategies and for understanding failures. Broadcast video providers, on the other hand, often need to automatically generate summaries and highlights of soccer matches, currently still achieved via manual annotation in most cases.

For these reasons, there has been an increasing effort to develop architectures for spotting actions in soccer videos~\cite{giancola2018soccernet,vats2020event,cioppa2020context}. The task requires to temporally localize all significant events happening inside the match like goals or yellow/red card events. As these events are sparse within a video, the task couples the need for effective feature extraction with that of properly handling the data imbalance and event sparsity issues. Also, an effective action spotting approach needs to provide accurate temporal localization, which in turn requires proper architectural and training strategies.

In this paper, we treat the action spotting as a detection problem and devise a novel network which takes inspiration from the regression strategies used in object detection. Specifically, our network takes a short video snippet as input and predicts the presence of a candidate event together with its class and relative temporal position inside the input snippet. This choice is innovative compared to recent works in the field, which usually assign the event to the central frame of the video chunk, thus preventing the model from learning an accurate localization of actions. We also deal with the sparsity and class imbalance issues through a sampling approach that ensures a uniform distribution in training batches in terms of ground-truth class and action locations.

Further, we develop a strategy to increase the generalization capabilities of the network, which is inspired by the masking strategies used in self-supervised pre-training techniques~\cite{devlin2018bert}. During training, indeed, we randomly mask and replace the frames preceding an event and constraint the network to focus on the most relevant parts of the action. 

\begin{figure}[t]
\centering
\includegraphics[width=1\linewidth]{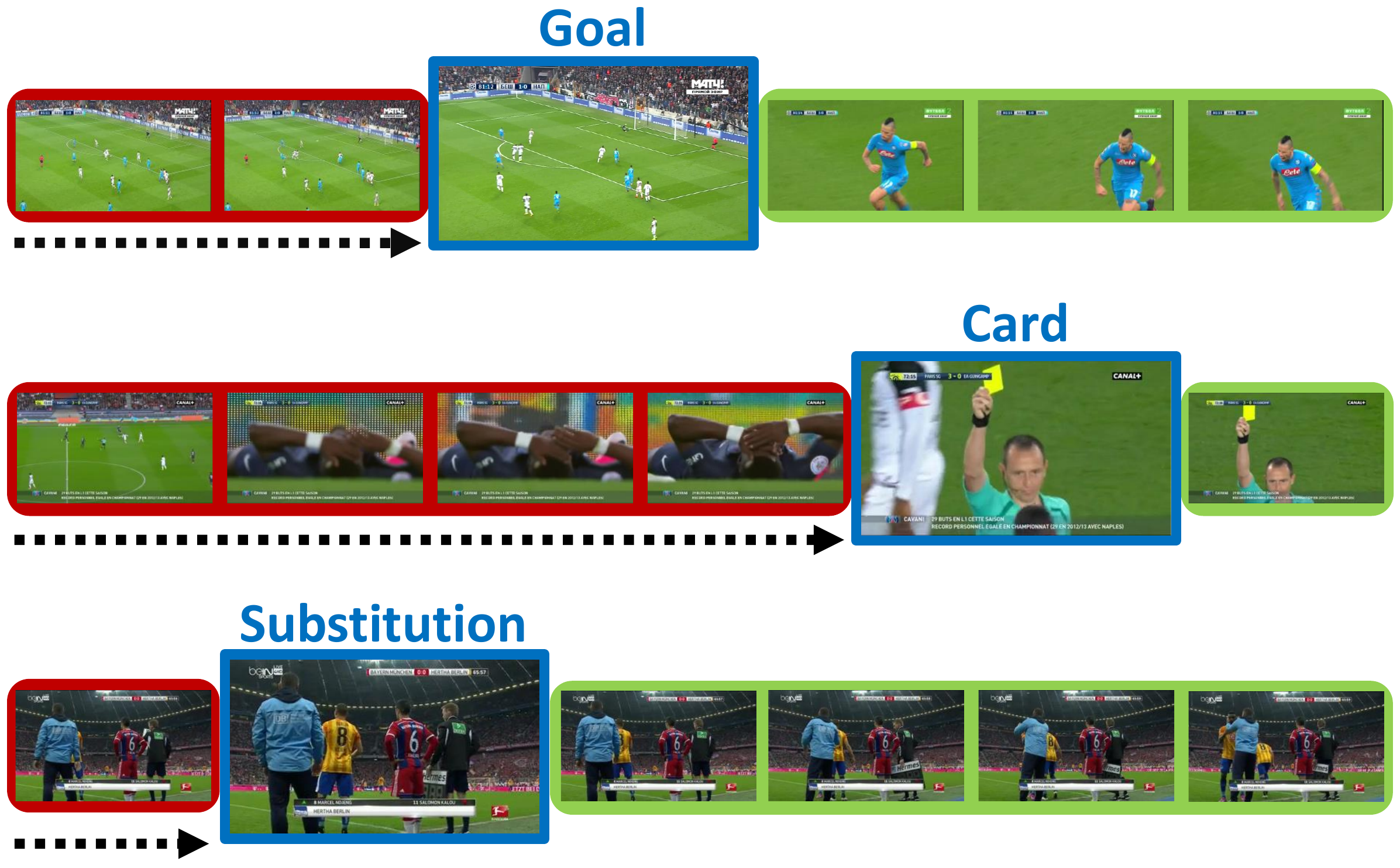}
\caption{We propose a lightweight and effective network for spotting relevant events in soccer matches. Our model features a masking strategy which increases training performance by constraining the network to focus on the most relevant parts of the video and a data sampling solution which handles class imbalance.}
\label{fig:first_page}
\end{figure}

We conduct extensive evaluations on the recently released SoccerNet dataset, which features 500 full broadcast soccer matches, annotated with relevant events. We provide experiments to validate the architectural choices behind the proposal and the effectiveness of the masking and data sampling strategies. When compared to the state of the art, our network achieves a gain of 3 Average-mAP points exploiting the same features used by previous works, while being significantly more lightweight. We also conduct experiments with different feature extraction backbones, and investigate the role of fine-tuning such backbones, devising a combination which further pushes the state of the art of 10 Average-mAP points.

\section{Related Work}

\tit{Video activity understanding}
Video Understanding is one of the most challenging and broad areas of Computer Vision, and many research efforts have been dedicated to this field in the last few years. As understanding the complexity of videos often requires to learn data-driven models with many parameters, the effort of researchers has also focused on the collection of proper datasets in recent times. Some of the datasets have collected action clips from movies or user-generated videos, like HMDB~\cite{kuehne2011hmdb}, UCF101~\cite{soomro2012ucf101}, YouTube-8m~\cite{abu2016youtube}, or Kinetics~\cite{kay2017kinetics}, while others have focused on more domain-specific actions, like those belonging to soccer events and sports in general~\cite{karpathy2014large,giancola2018soccernet}.

On the technical and architectural side, most of the recent methods employ variants of temporal convolutions to integrate motion and appearance features, either by using full 3D kernels~\cite{tran2015learning, carreira2017quo} or separating the extraction of spatial and temporal features in different layers~\cite{feichtenhofer2016spatiotemporal, varol2017long}. Other techniques integrate motion in a second stream of the architecture which processes the optical flow of the video~\cite{feichtenhofer2016convolutional,simonyan2014two}. Recently, the literature has also concentrated on the role of the spatial and temporal resolution of the video~\cite{feichtenhofer2019slowfast, Wu_2020_CVPR}.

Besides classifying the actions performed inside a given video, the literature has also investigated the prediction of the temporal boundaries of the action~\cite{zeng2019graph, lin2019bmn, lin2018bsn}, with datasets such as ActivityNet~\cite{caba2015activitynet} or Thumos~\cite{idrees2017thumos}, and on the spatio-temporal location of actions~\cite{gu2018ava}.

\begin{figure}[t]
\centering
\includegraphics[width=1\linewidth]{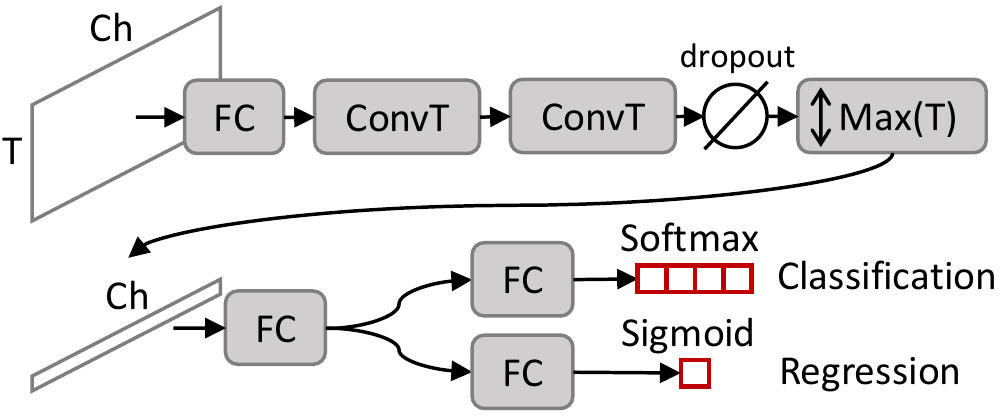}
\caption{Architecture of \OURS.}
\label{fig:model}
\end{figure}

\tit{Soccer event spotting}
On a related line of research, the action spotting task has been proposed for soccer videos along with the SoccerNet dataset in~\cite{giancola2018soccernet}, with the aim of finding the exact anchor time (or \textit{spot}) of an event and to recognize it.

The final objective of these approaches is primarily highlights identification and generation. With this purpose, Bag-of-Words and SIFT features have been adopted by~\cite{baccouche2010action}, together with an LSTM for soccer video classification, while deep convolutional features have been used in~\cite{jiang2016automatic}. Tsagkatakis~\etal~\cite{tsagkatakis2017goal} adopted an optical flow and appearance feature fusion strategy for \textit{goal} and \textit{no goal} event detection. After the spotting problem definition by Giancola~\etal~\cite{giancola2018soccernet} and a first baseline, several approaches have been proposed for this task: audio stream integration has been explored in~\cite{vanderplaetse2020improved}, showing promising spotting results. Vats~\etal~\cite{vats2020event} introduced a multi-tower temporal convolutional network, while a context-aware loss function has been defined in~\cite{cioppa2020context}, observing that frames \textit{just after} an event contain most of the visual cues for event recognition.

\tit{Large-scale Soccer datasets}
Gathering a large number of realistic soccer videos is not easy, because of the limited public availability of broadcast content. Nevertheless, a number of datasets for soccer analysis have been recently proposed. SoccerNet~\cite{giancola2018soccernet} is a collection of 500 broadcast soccer games with one second resolution event annotations, and is at present the largest dataset for soccer action spotting. Yu~\etal~\cite{yu2018comprehensive} collected 222 soccer matches with shot transitions, event boundaries, and players bounding box annotations. SoccerDB~\cite{wang2019comprehensive} includes 346 soccer games with event segments and players, ball, goalposts bounding box annotations.

\section{Proposed Method}
In the following, we present our approach for soccer event spotting. Our formulation features a lightweight network, which can be integrated with any existing backbone, and which jointly predicts classification scores and temporal offsets. This is combined with a masking strategy which increases the training performance, and with effective handling of data imbalance.

\subsection{Proposed architecture}

A soccer match can be represented as a sequence of frames $\left( x_{1}, x_{2}, ..., x_{N} \right)$, with $x_{i} \in \mathbb{R}^{H \times W \times Ch}$, extracted with a given frame rate from the original video clip. The goal of our approach is to spot and classify a set of interesting events inside the match, \ie~to predict a set of pairs $\{(t_{j}, e_{j})\}_{j=1}^n$, where $t_j$ indicates the timestamp of the action from the beginning of the video, and $e_j \in \{0, ..., C-1\}$ indicates its class label.


Taking inspiration from the regression strategies used in object detection~\cite{ren2015faster}, our model takes as input a short video chunk $\bm{X} = \left(x_{1}, x_{2}, ..., x_{T} \right)$ with length $T$, and predicts the temporal offset of a possible action inside the chunk, plus a classification score over $C+1$ classes, where the additional class indicates the background one. At test time, predictions can be obtained by applying the model on chunks extracted from the input video with a given stride, converting relative offsets to absolute timestamps, and accumulating predictions over time.

\label{sub:masking}
\begin{figure}[t]
\centering
\includegraphics[width=1\linewidth]{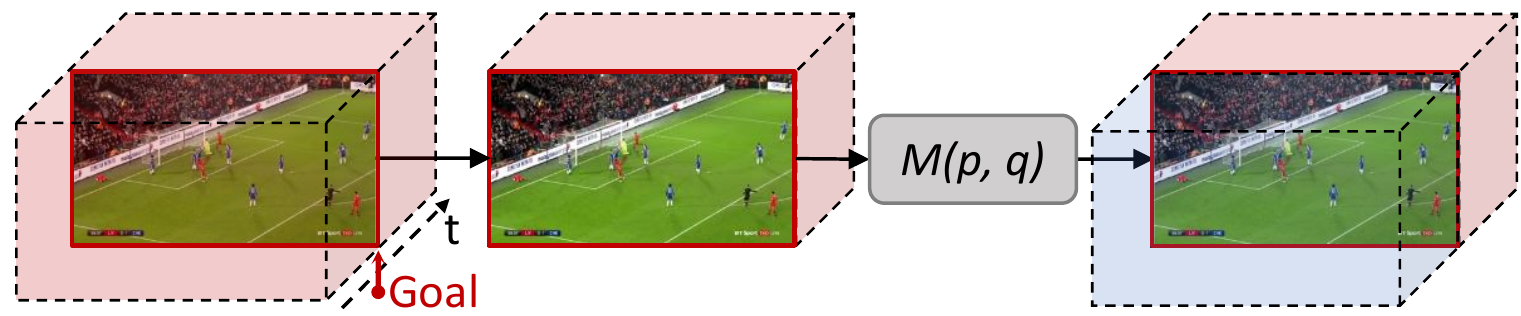}
\caption{Masking strategy. Frames before the event are masked with probability \textit{p}, if the event lies before the offset \textit{q}.}
\label{fig:masking}
\end{figure}

In our network, the $T$ input frames are independently passed to a 2D backbone to extract a feature vector for each frame, which is obtained by averaging the activation map's spatial dimensions from the last convolutional layer and linearly projecting the result to a common dimensionality. The model then applies a stack of two 1D convolutions over the temporal dimension to combine the features of different frames. After the convolutions, a \textit{maximum} operation is applied to remove the time axis, and a stack of linear layers is added. The tail of the model consists of two sibling fully-connected layers, one for action classification and the other for temporal offset regression. 
Our model is visually depicted in Fig.~\ref{fig:model}.

Given a ``foreground'' input clip containing an event $(t_j, e_j)$, the classification layer outputs per-class probabilities and a cross-entropy loss is applied:
\begin{equation}
\label{eq:xe}
\mathcal{L}_{cls} = -\sum_{c=0}^{C}\mathds{1}_{c = e_j} \log(p_{c}),
\end{equation}
where $C+1$ is the number of different event labels, including the \textit{background} label, $\mathds{1}_{c = e_j}$ is the indicator function, and $p_{c}$ is the model output probability for event label $c$.
On the other hand, the regression layer predicts a single scalar per clip, which is projected in the range $[0, 1]$ through a sigmoid function and trained to predict the normalized relative offset of the event in the video chunk. Formally, we apply a squared-error loss as follows:
\begin{equation}
\label{eq:mse}
\mathcal{L}_{regr} = (\sigma(o) - r_j)^2,
\end{equation}
where $o$ is the output of the regression branch of the network, and the normalized relative offset $r_j$ is computed as $(t_j - s)/T$, where $s$ is the starting timestamp of the video chunk.

The final loss for a foreground chunk is a weighted sum of $\mathcal{L}_{cls}$ and $\mathcal{L}_{regr}$:
\begin{equation}
\label{eq:loss}
\mathcal{L} = \mathcal{L}_{cls} + \lambda\mathcal{L}_{regr};
\end{equation}
for ``background'' video chunks, \ie~chunks which do not contain any event, we instead apply only the classification loss, using the background class as ground truth label.

As it can be noticed, the proposed method predicts a single event per clip. It could be easily extended to predict more than one event, \eg~including for instance a bipartite matching strategy between predictions and ground truth labels~\cite{carion2020end}. In practice, we noticed that interesting events are very sparse in a soccer match. The only rare circumstances in which more than one action occurs in a few seconds interval correspond to  double substitutions or double yellow/red cards. While the choice of having a single action predictor could slightly affect spotting performances, it does not affect automatic highlight generation, which is the final goal of event spotting in soccer videos and allows us to maintain a simple and lightweight architecture.

\subsection{Masking strategy}

The majority of visual cues that contribute to the recognition of an event occur \textit{just after} the event itself~\cite{cioppa2020context}. This is particularly evident in the case of soccer matches, in which reactions to an event are often a good indicator of the presence of the event itself, like in the case of the celebrations following a \textit{goal} event. Taking inspiration from the masking strategies used in self-supervised pre-training approaches~\cite{devlin2018bert}, we endow our approach with a masking policy that allows the network to focus on the most relevant portions of the clips at training time.

In our formulation, a foreground training clip is masked with a given probability by replacing the frames before the event with a randomly chosen background sequence (\ie~which does not contain any event). We also make sure that there is a sufficient number of frames after the event, by avoiding the application of masking on clips in which the event comes too late. Background training clips, instead, are not masked at all.

Formally, we define a masking function $M(p,q)$, in which $p$ indicates the masking probability and $q$ is the maximum normalized relative temporal offset of the event inside the clip. Given a foreground clip $\bm{X}$ containing an event $(t_j, e_j)$, the masking function $M(p,q)$ is defined as follows:
\begin{equation}
\label{eq:masking}
M(p,q)(\bm{X}) = 
\begin{cases}
    \left(z_{1}, ..., z_{t_j-s-1}, x_{t_j-s}, ..., x_T\right) \\ \quad \text{if } r_j \leq q \text{, } u<p\\ \\
    \left(x_{1}, ..., x_{t_j-s-1}, x_{t_j-s}, ..., x_T\right) \\ \quad \text{otherwise},
\end{cases}
\end{equation}
where $s$ is the starting timestamp of the video chunk, $r_j$ is the normalized relative offset of the event inside the video chunk, $(z_i)_{i=1}^{t_j-s-1}$ is a sequence of consecutive frames randomly selected from a background clip, and $u$ is a random value sampled from the uniform distribution $U \left[0,1\right]$. A visualization of the masking strategy is also reported in Fig.~\ref{fig:masking}.

Through this masking approach, we randomly force the model to recognize an event using just the frames \textit{following} the event itself, and without relying on the previous ones. We will experimentally show how this masking strategy, which does not require any architectural change or additional resources, is robust to different values of the masking probability $p$, and how it can increase the recognition and localization accuracy of the model. 

\subsection{Data Sampling and Balancing}
\label{sub:augm}
The training set of our architecture is composed of all possible clips with length $T$ which can be extracted from a set of matches. Since relevant events in a soccer match are quite sparse, a balancing strategy is needed to make sure that the network can learn to properly classify relevant events, without losing the capability of distinguishing background clips. At the same time, we need to make sure that the distribution of the offsets used to train the regression branch is sufficiently uniform.

Given a training soccer match and an anchor event $(t_{j}, e_{j})$, we extract all clips with length $T$ containing the event, sliding a window along the time axis with stride $1$. Doing so, we generate many clips for the same event, where each of them contains the event in a different relative temporal location. Repeating this procedure for each event in a match, and then for each match in the training set, we collect a set of interesting events, where the distribution of the relative position of events inside clips is balanced by construction.

The remaining parts of the matches, which do not contain any event, are sliced with a window of size $T$ and stride $T$ (thus avoiding overlapping), to build the set of background video chunks.
In each training epoch, we randomly sample $n_{F}$ foreground clips from the above mentioned set, and $n_{B} = n_{F} / C$ clips (where $C$ is the number of classes) from the set of background clips, to balance the number of samples per class. During inference we extract non-overlapping clips, each with $T$ frames, in a sliding-window manner, assuming that $T$ is small enough to prevent that more than one action occurs inside a clip.

\section{Experimental Evaluation}
\label{sec:experiments}

\subsection{Dataset and evaluation protocol}
\tit{Data}
We train and evaluate the proposed method on the SoccerNet dataset~\cite{giancola2018soccernet}. SoccerNet gathers 500 full broadcast soccer matches, spanning 764 hours of video, which are split into 300 games for training, 100 for validation, and 100 for testing. Interesting events belong to three categories (\textit{goal}, \textit{card}, \textit{substitution}) and are manually annotated with a temporal resolution of one second. The average separation between events is 6.9 minutes, thus leading to a very sparse annotation. 

For fairness of comparison with previously published baselines and state-of-the-art approaches, all experiments are performed with the pre-computed ResNet-152~\cite{he2016deep} features released with the dataset itself, unless otherwise specified. These have been obtained by Giancola~\etal~\cite{giancola2018soccernet} by resizing and cropping videos at a resolution of $224 \times 224$ and extracting frames at a frequency of 25 fps. Starting from this extraction, a feature vector was computed every 0.5 seconds. They also applied a PCA step to reduce the dimensionality to 512. The feature extraction backbone was pre-trained on ImageNet~\cite{deng2009imagenet}.

\tit{Evaluation metric}
The action spotting task requires to correctly predict the anchor \textit{spot} that identifies an event. For instance, a ground truth spot for a \textit{card} event is defined as the timestamp in which the referee extracts the card. Following previous literature, we use the Average-mAP defined in~\cite{giancola2018soccernet} as the evaluation metric, which accounts for multiple temporal tolerances. 
Given a temporal tolerance $\delta$, we compute the average precision for a class by considering a prediction as positive if the distance from its closest ground truth spot is less than $\delta$. The mean average precision is then obtained by averaging the AP of each class. The final metric is computed as the area under the mAP curve obtained by varying $\delta$ in the interval ranging from 5 to 60 seconds.
 
\begin{table}[t]
\centering
\caption{Structure of the proposed architecture.}
\begin{tabular}{ccc}
\toprule 
Layer & Input Channels & Output Channels \\
\midrule
$\text{FC}_1$ & 512 & 256 \\
$\text{Conv}_1$ & $9 \times 256$ & 256 \\
$\text{Conv}_1$ & $9 \times 256$ & 128 \\
$\text{Dropout}$ & -- & -- \\
$\text{Max over time}$ & -- & -- \\
$\text{FC}_2$ & 128 & 64 \\
$\text{FC}_{cls}$ & 64 & $C$ (number of classes) \\
$\text{FC}_{regr}$ & 64 & 1 \\
\bottomrule
\end{tabular}
\label{tab:channels}
\end{table}

\begin{table*}[t]
\centering
\caption{Comparison with baselines and state-of-the-art approaches.}
\begin{tabular}{lcccc}
\toprule 
Model & Clip length (s) & Features & Val Avg-mAP & Test Avg-mAP \\
\midrule
SoccerNet baseline~\cite{giancola2018soccernet} & 5 & ResNet-152 (PCA) & - & 34.5 \\
SoccerNet baseline~\cite{giancola2018soccernet} & 60 & ResNet-152 (PCA) & - & 40.6 \\
SoccerNet baseline~\cite{giancola2018soccernet} & 20 & ResNet-152 (PCA) & - & 49.7 \\
Vanderplaetse~\etal~\cite{vanderplaetse2020improved} & 20 & ResNet-152 (PCA) + Audio & - & 56.0 \\
Vats~\etal~\cite{vats2020event} & 15 & ResNet-152 (PCA) & - & 60.1 \\
Cioppa~\etal~\cite{cioppa2020context} & 120 & ResNet-152 (PCA) & - & 62.5 \\
\midrule
Ours & 20 & ResNet-152 (PCA) & \textbf{67.8} & \textbf{65.5} \\
\bottomrule
\end{tabular}
\label{tab:sota}
\end{table*}

\subsection{Implementation details}
\label{sub:impl_details}
In all our experiments, the model takes $T=41$ frames as input, therefore spanning 20 seconds of the match. The weight $\lambda$ of the regression loss (defined in Eq. ~\ref{eq:loss}) is set to 10, unless otherwise specified. Table~\ref{tab:channels} reports the architectural details of the model, including the number of input and output channels. Each 1D convolutional layer has a kernel size of 9, a stride of 1, and zero-padding to keep the temporal dimension shape. The drop rate of the dropout layer is set to 0.1.

\tit{Training}
Following our data sampling strategy, we obtain a total of around 150,000 foreground and 72,000 background video chunks.
At each training epoch, we randomly sample 30,000 foreground sequences and 10,000 background sequences, being the number of foreground classes $C$ equal to three in SoccerNet. During training, we also drop all \textit{substitution} events occurring at half time, since no visual cues suggest that a substitution is happening. 

The masking probability $p$ is set to $1/3$, while the maximum relative temporal offset $q$ is set to 0.5, unless otherwise specified. 
We train our model for a maximum of 50 epochs using an SGD optimizer with momentum 0.9. The batch size is set to 64 and we apply a learning rate of 0.05, with a linear warm-up during the first epoch and a cosine annealing scheduling from the second epoch. A weight decay of $10^{-4}$ is also adopted. Early stopping on the Average-mAP computed over the validation set is applied. Training is done on a NVIDIA RTX 2080Ti GPU.

\tit{Inference}
During inference, we slide a window containing $T$ frames on the input video with a stride of $T$, and the clip class, together with the event relative temporal offset, is predicted by the two sibling fully connected layers. The predicted relative offset is then converted to absolute timestamp, to obtain the predicted spot location. No masking is applied during inference.

\subsection{Main results and comparison with the state of the art}
In the following, we validate the proposed approach for action spotting in soccer video. We firstly perform a comparison with baselines and state-of-the-art methods and then analyze and ablate the key components of the approach. Finally, we will investigate the performance of the proposed architecture when using different 2D convolutional backbones.

\tit{Comparison with baselines and previous methods}
In Table~\ref{tab:sota} we report the performance of our model, in comparison with previous approaches for action spotting. All the reported approaches adopt the ResNet-152 features released with the dataset, and we do the same for fairness of comparison. The only exception is the approach presented by Vanderplaetse~\etal~\cite{vanderplaetse2020improved}, which enriched the same visual features by encoding the audio stream through a VGG network~\cite{simonyan2014very}.

As it can be noticed, our approach outperforms the best SoccerNet baseline (with $T=20$ seconds) by 15.8 Average-mAP points and the to-date best performing method by 3 Average-mAP points on the test set of SoccerNet. Noticeably, this result is achieved without any matching strategy between predictions and ground truth, and without post-processing steps like non-maxima-suppression. 

We also conduct a grid search to find the optimal clip duration $T$, as done in Giancola~\etal~\cite{giancola2018soccernet} on the SoccerNet baselines (and reported in Table~\ref{tab:sota}). Also with our architecture, the optimal clip duration is 20 seconds.

\tit{Per-class performances}
\begin{figure}[t]
\centering
\includegraphics[width=0.8\linewidth]{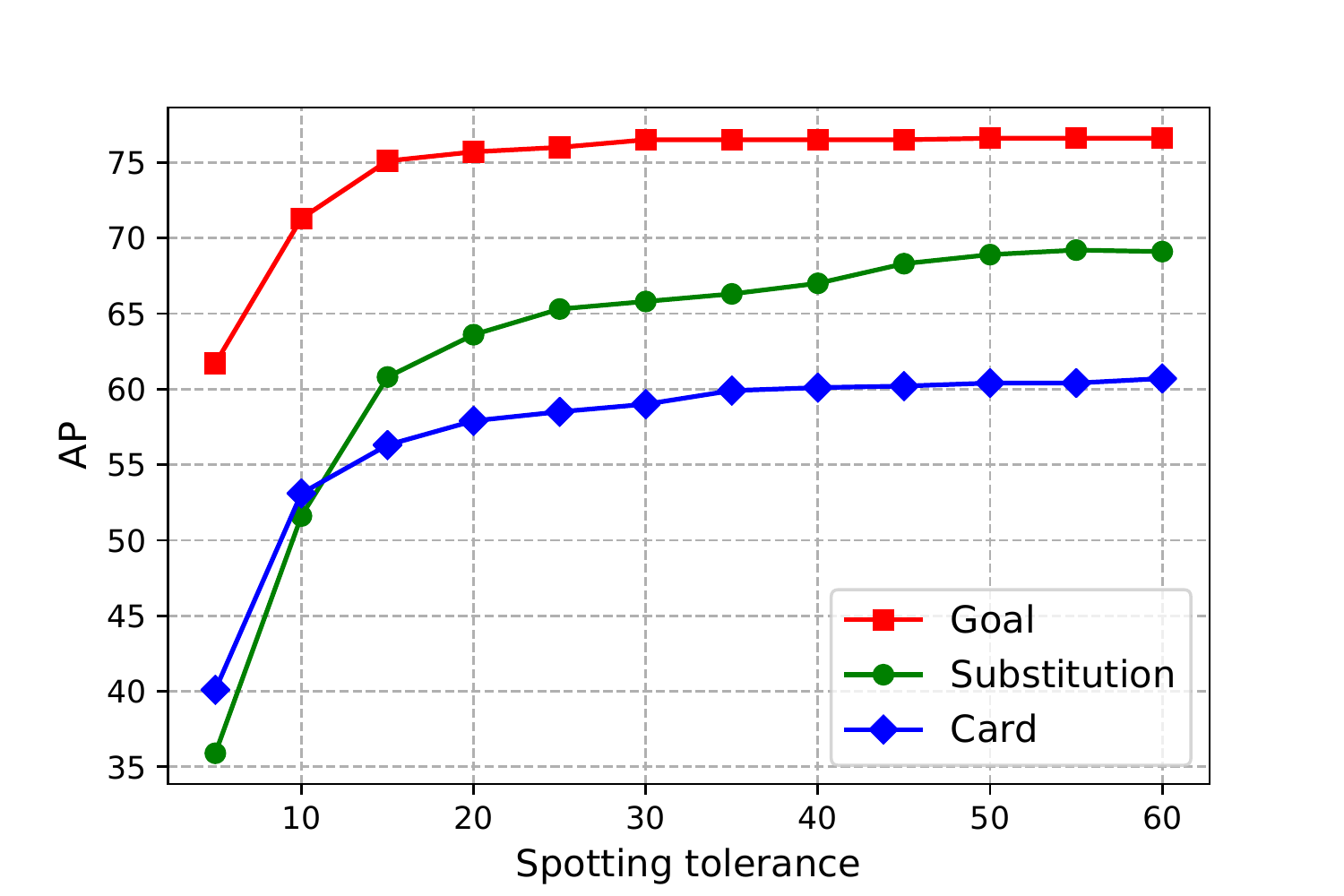}
\caption{Per-class Average Precision, as a function of spotting tolerance.}
\label{fig:mAP_per_class}
\end{figure}
In Fig.~\ref{fig:mAP_per_class} we report the AP computed for each class, as a function of the spotting tolerance. The best performing events are \textit{goals}, with a margin of around 5 AP points compared to the \textit{substitution} class when allowing high spotting tolerances. This is in line with previous literature, and can be attributed to the richness of visual cues which are usually found after a \textit{goal} event (\eg~celebration, replays). \textit{Card} events are, instead, the most difficult to spot, as the presence of a yellow or red card is the only visual indication that can distinguish those events from a general foul. Finally, it can be observed that \textit{goals} are also the events which can be localized with the greatest temporal precision, compared to the other classes.

\subsection{Ablation studies}
\label{sec:ablations}
\tit{Removing key components}
We investigate the role of the key components of our approach by conducting an ablation study. 
Firstly, we assess the role of employing a uniformly distributed normalized relative offset in foreground clips. To this aim, we modify our data sampling strategy by considering only the chunks that contain an event in the middle of the clip (thus, with a normalized relative offset of 0.5), and removing the regression output. In this case, we still balance training batches by ensuring a uniform number of clips for each class, including background. At prediction time, we assume a normalized relative offset of 0.5. Noticeably, this evaluation setting is similar to the one adopted in~\cite{giancola2018soccernet} and~\cite{vats2020event}. As can be seen from Table~\ref{tab:abl}, this leads to a significant performance drop, thus testifying the need for uniformly distributing relative offsets. 

\begin{figure}[t]
\centering
\includegraphics[width=0.8\linewidth]{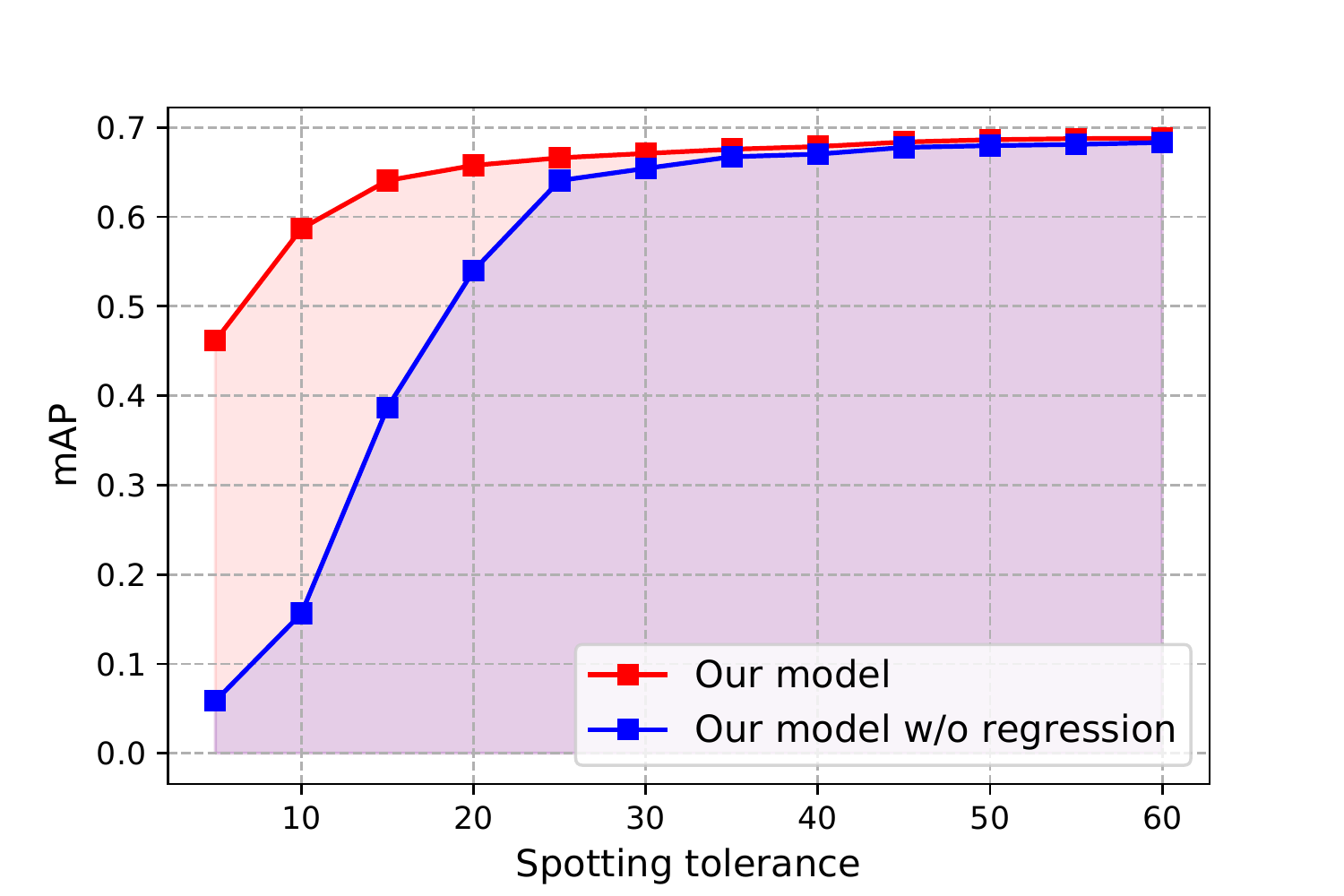}
\caption{Mean average precision when varying the tolerance $\delta$ in the interval between 5 and 60 seconds, for our complete model (red curve) and when removing the offset regression branch (blue curve). The Average-mAP gain can be quantified as the area under the red curve and above the blue one.}
\label{fig:mAP}
\end{figure}

In a second ablation experiment, we instead maintain the original data sampling strategy and exclude the regression branch, to estimate its contribution to the final performance. Also in this case, we assume that the relative temporal offset of a predicted event is always $0.5$. This leads to a drop of the test Average-mAP of almost 10 points, as reported in Table~\ref{tab:abl}. Fig.~\ref{fig:mAP} also shows a more detailed comparison for different values of the tolerance threshold $\delta$ using this regression-free model (in blue) and our full proposal (in red). We notice that, for high values of $\delta$, the resulting mAP is similar for both models. When $\delta$ decreases (and in particular when $\delta$ is lower than the clip duration) our model avoids an abrupt decrease of the mAP, which instead occurs when removing the regression branch. This confirms that the regression loss can increase the localization accuracy of the prediction.

Further, we also test the contribution given by the masking strategy. When removing the masking, we observe a decrease of 1.5 Average-mAP points on the test set, as reported in Table~\ref{tab:abl}. We also underline how masking represents a cost-free improvement, which does not require any additional resource or model parameter.

\begin{table}[t]
\centering
\caption{Performance of the proposed model when removing key components.}
\begin{tabular}{ccc}
    \toprule 
    Model & Val Avg-mAP & Test Avg-mAP \\
    \midrule
    Ours & \textbf{67.8} & \textbf{65.5} \\
    Ours w/o uniformly distributed offsets & 48.7 & 46.2 \\
    Ours w/o offset regression branch & 58.5 & 55.7 \\
    Ours w/o masking & 66.5 & 64.0 \\
    \bottomrule
\end{tabular}
\label{tab:abl}
\end{table}

\tit{Masking analysis}
In Table~\ref{tab:mask_bafore} we show how performances vary when changing the masking probability $p$ and the  maximum relative temporal offset $q$ for masking. We achieve the best configuration with $p=1/3$ and $q=0.5$, which means that clips with the event in their first half have a $1/3$ probability of being masked during training. Keeping $q=0.5$ fixed and varying $p$, the Average-mAP always exceeds the performance of the masking-free model (64.0 Avg-mAP) except for $p=1$. When $p=1$, indeed, a clip is always masked if it has the event in the first half. This creates a big gap between the train and test distributions and performances drop as expected. Similarly, the model is robust to different values of $q$, except when all the clips have probability $p=1/3$ to be masked, independently of where the event lies inside it (\ie~when $q=1$). In this case, clips with too little context after the event can be masked too, resulting in lower Average-mAP.

One can question if the majority of visual cues actually occurs just after the event, and not before. Table~\ref{tab:mask_after} shows the results when masking frames \textit{just after} the event with different values of $p$, while always keeping the frames \textit{before} the event. Under this setting, masking is not beneficial and lowers the final performance.

\begin{table}[t]
\centering
\caption{Performance when varying the masking probability $p$ and the maximum relative temporal offset $q$ for masking.}
\begin{tabular}{ccccc}
    \toprule 
    $p$ & $q$ & & Val Avg-mAP & Test Avg-mAP \\
    \midrule
    1/5 & 0.5 & & 66.8 & 64.6 \\
    1/4 & 0.5 & & 67.0 & 64.4 \\
    1/3 & 0.5 & & \textbf{67.8} & \textbf{65.5} \\
    1/2 & 0.5 & & 67.1 & 64.4 \\
    1 & 0.5 & & 64.7 & 60.7 \\
    \cmidrule{1-2}
    \cmidrule{4-5}
    1/3 & 0.1 & & 65.5 & 63.4 \\
    1/3 & 0.25 & & 67.4 & 64.7 \\
    1/3 & 0.5 & & \textbf{67.8} & \textbf{65.5} \\
    1/3 & 0.75 & & 66.5 & 64.0 \\
    1/3 & 1 & & 64.7 & 62.6 \\
    \bottomrule
\end{tabular}
\label{tab:mask_bafore}
\end{table}

\begin{table}[t]
\centering
\caption{Performance when varying $p$, keeping $q=0.5$ fixed and masking frames \textit{after} the event.}
\begin{tabular}{ccccc}
    \toprule 
    $p$ & $q$ & & Val Avg-mAP & Test Avg-mAP \\
    \midrule
    1/5 & 0.5 & & 65.2 & 62.5 \\
    1/4 & 0.5 & & 64.6 & 62.9 \\
    1/3 & 0.5 & & 63.8 & 61.8 \\
    1/2 & 0.5 & & 61.4 & 60.7 \\
    1 & 0.5 & & 54.1 & 54.1 \\
    \bottomrule
\end{tabular}
\label{tab:mask_after}
\end{table}

\tit{Additional analysis}
\begin{figure}[t]
\centering
\includegraphics[width=0.8\linewidth]{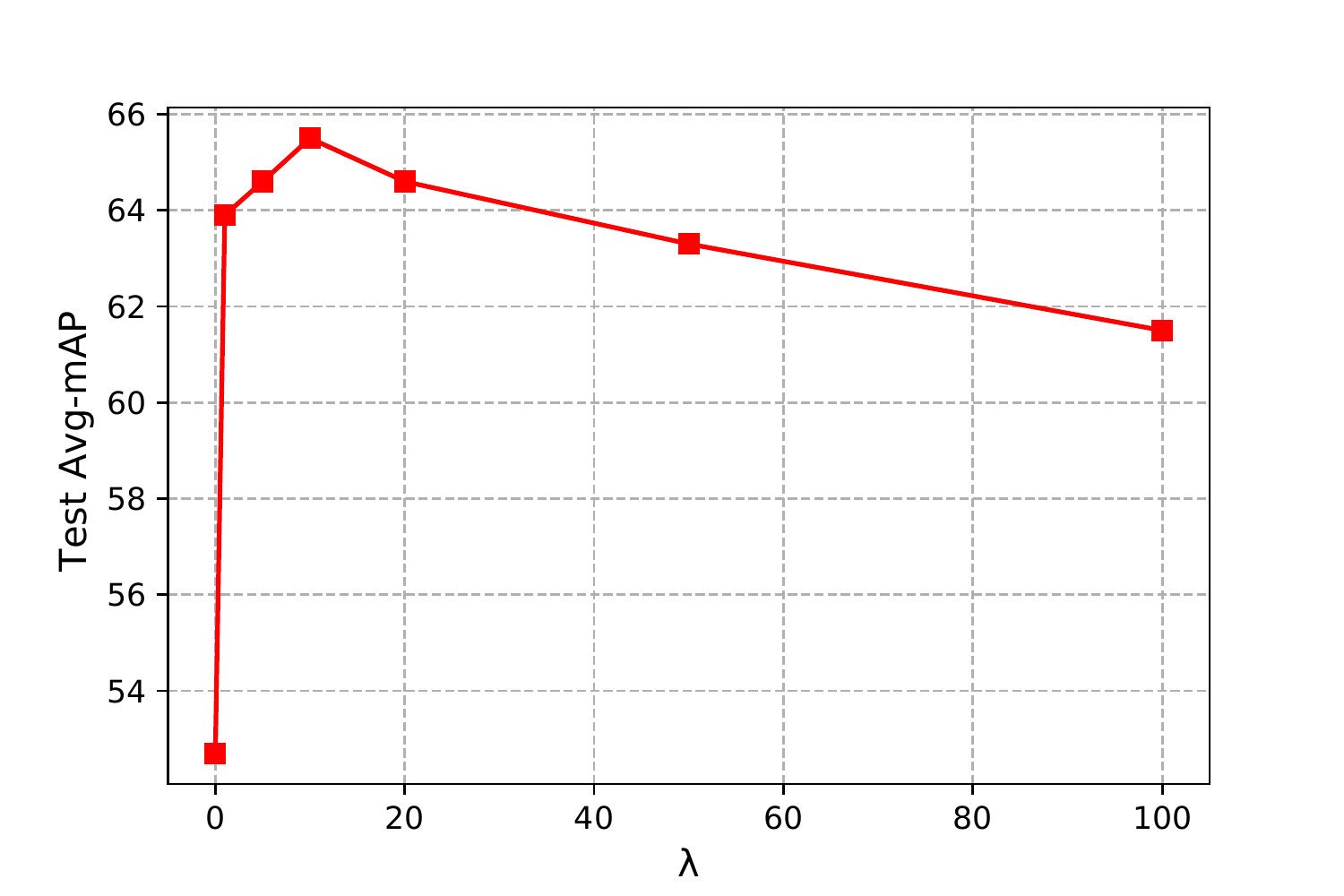}
\caption{Performance when varying the regression weight $\lambda$.}
\label{fig:lambda}
\end{figure}
Finally, we look for the best value of hyperparameter $\lambda$, which controls the relative weight of the regression loss with respect to the classification loss (see Eq.~\ref{eq:loss}). Fig.~\ref{fig:lambda} shows the test Average-mAP when varying $\lambda$ in the interval between 0 and 100. The best value is achieved with $\lambda=10$. When $\lambda=0$ there is no temporal offset regression in training, and we fall back in a setting similar to that of Table~\ref{tab:abl}, third row. In this case, however, the relative temporal offset is still predicted by the network (without any supervision) and not fixed to the clip center timestamp. The gap between the setting with $\lambda=0$ and $\lambda=1$ exceeds 10 Average-mAP points. Performances decrease, instead, when $\lambda>10$.

\subsection{Changing the convolutional backbone}
Further, we present additional results when changing the convolutional backbone used for feature extraction, and when fine-tuning part of it during the training of the action spotting model.
Here, we employ different variants of ResNet, pre-trained on ImageNet. For assessing the role of fine-tuning, only the parameters belonging to the last residual block and our model are trained, while the other parameters of the backbone are kept fixed, as we did not observe significant improvements when fine-tuning larger portions of the backbone.

Table~\ref{tab:e2e} shows the performances obtained when finetuning ResNet-18, ResNet-50 and ResNet-152. We train these models on two RTX 2080Ti GPUs, with a batch size of 24 clips and a base learning rate of 0.025. RGB frames are extracted from the low resolution videos ($224 \times 398$) at a rate of 2 fps, thus keeping the entire frame instead of center cropping as done in the features released with SoccerNet. No spatial augmentation is performed, and all the other implementation details are kept unchanged.

When using ResNet-152 as our backbone, we observe a boost of 9.6 Average-mAP points on the test set when compared to our best performing model trained on the pre-computed ResNet-152 features. A similar performance level is obtained when finetuning ResNet-50, while a more significant performance loss is visible when using ResNet-18. While pre-computed features are a good starting point for research and comparison purposes, our findings underline that end-to-end training still guarantees a significant performance boosting.

\begin{table}[t]
\centering
\caption{Performance analysis when finetuning different variants of ResNet.}
\begin{tabular}{cccc}
    \toprule 
    Model & Pre-train & Val Avg-mAP & Test Avg-mAP \\
    \midrule
    ResNet-18 + Our & ImageNet & 73.8 & 70.9 \\
    ResNet-50 + Our & ImageNet & 76.6 & 74.9 \\
    ResNet-152 + Our & ImageNet & \textbf{77.5} & \textbf{75.1} \\
    \bottomrule
\end{tabular}
\label{tab:e2e}
\end{table}

\subsection{Qualitative analysis}
\begin{figure}[t]
\centering
\includegraphics[width=1\linewidth]{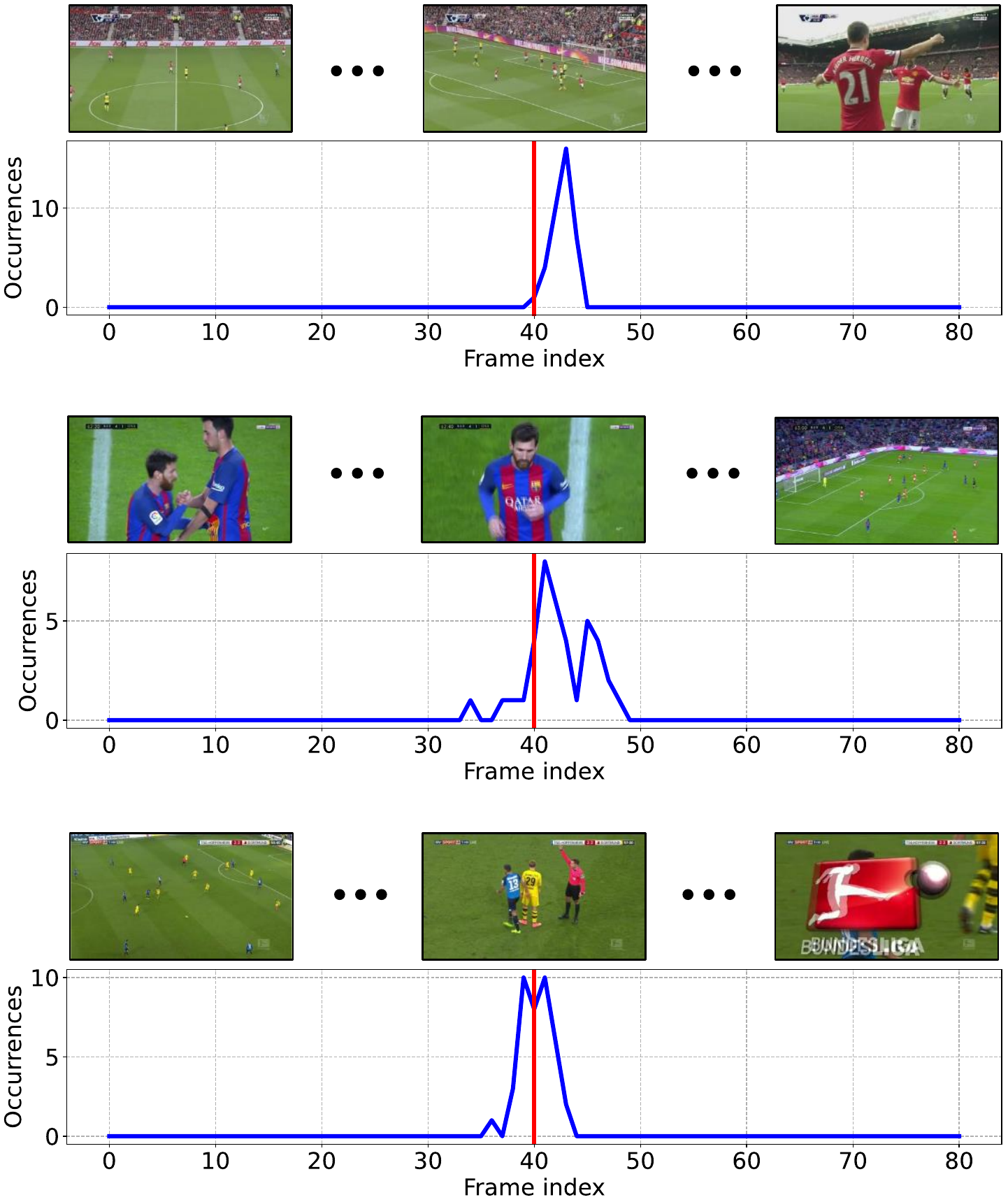}
\caption{Qualitative results. The ground truth action timestamp is shown in red, while the blue curve shows the number of times a frame index was predicted as spot. A \textit{goal}, a \textit{substitution} and a \textit{card} events are shown, from top to bottom.}
\label{fig:qualitatives}
\end{figure}

Finally, we qualitatively assess the spotting capabilities of our model. For this purpose, we extract frames from the raw video with a resolution of 2 fps and create overlapping clips having a length of $T$ frames, with stride 1.
For each clip, our model predicts the event temporal offset and its label, and we count the number of times a frame is predicted as a spot. As can be seen from Fig.~\ref{fig:qualitatives}, the most voted frame index (the peak of the blue plot) is often very close to the ground truth spot (highlighted in red), and usually lies in an interval of 10 frames around it, corresponding to the lowest tolerance value $\delta$ which is considered for the Average-mAP computation.

\section{Conclusion}
In this paper, we have presented a novel approach for spotting relevant actions inside soccer matches. Our proposal is a lightweight network including a simple stack of temporal operations after the extraction of frame-level features, and which jointly predicts classification scores and temporal offsets. The training of the network is paired with a masking strategy, which constraints the network to focus on the most relevant regions of the input chunks. Further, we devised an effective data sampling and balancing strategy which increases the recognition performance. The proposed approach has been extensively tested on SoccerNet, demonstrating the appropriateness of the components and the design choices. Our full model surpasses the state of the art by a significant margin when using the same features. Finally, we have conducted an experimental evaluation on the use of different backbones, which leads to an increase of about 10 Average-mAP points over the state of the art.

\section*{Acknowledgments}
This work has been partially funded by Metaliquid Srl, Milano (Italy). We also acknowledge NVIDIA AI Technology Center for providing technical support and computational resources used in this research.





\bibliographystyle{IEEEtran}
\balance
\bibliography{egbib}
%



\end{document}